\documentclass[conference]{IEEEtran}
\IEEEoverridecommandlockouts
\usepackage{cite}
\usepackage{amsmath,amssymb,amsfonts}
\usepackage{algorithmic}
\usepackage{graphicx}
\usepackage{textcomp}
\usepackage{hyperref}
\usepackage{flushend}
\usepackage[utf8]{inputenc}
\usepackage[russian, english]{babel}
\def\BibTeX{{\rm B\kern-.05em{\sc i\kern-.025em b}\kern-.08em
    T\kern-.1667em\lower.7ex\hbox{E}\kern-.125emX}}
    
\makeatletter
\def\ps@IEEEtitlepagestyle{%
  \def\@oddfoot{\mycopyrightnotice}%
  \def\@evenfoot{}%
}
\def\mycopyrightnotice{%
  {\footnotesize 978-0-7381-3111-5/20/\$31.00 \textcopyright2020 IEEE\hfill}
  \gdef\mycopyrightnotice{}
}    
    
\makeatletter
\def\footnoterule{\relax%
  \kern-5pt
  \hbox to \columnwidth{\vrule width 1.0\columnwidth height 0.4pt}
  \kern4.6pt}
\makeatother
    
\begin{document}


\title{Entity Recognition and Relation Extraction\\ 
from Scientific and Technical Texts in Russian\\

\thanks{The study was funded by Russian Foundation for Basic Research according to the research project N 19-07-01134.}

}

\author{\IEEEauthorblockN{Elena Bruches}
\IEEEauthorblockA{
\textit{A.P. Ershov Institute of Informatics Systems SB RAS} \\
\textit{Novosibirsk State University}\\
Novosibirsk, Russia \\
bruches@bk.ru} \\
\IEEEauthorblockN{Tatiana Batura}
\IEEEauthorblockA{
\textit{A.P. Ershov Institute of Informatics Systems SB RAS} \\
\textit{Novosibirsk State University}\\
Novosibirsk, Russia \\
tatiana.v.batura@gmail.com} \\
\and
\IEEEauthorblockN{Alexey Pauls}
\IEEEauthorblockA{
\textit{Novosibirsk State University} \\
Novosibirsk, Russia \\
aleksey.pauls@mail.ru} \\
\\
\IEEEauthorblockN{Vladimir Isachenko}
\IEEEauthorblockA{
\textit{A.P. Ershov Institute of Informatics Systems SB RAS} \\
Novosibirsk, Russia \\
vv.isachenko@gmail.com}
}

\maketitle

\begin{abstract}
This paper is devoted to the study of methods for information extraction (entity recognition and relation classification) from scientific texts on information technology. Scientific publications provide valuable information into cutting-edge scientific advances, but efficient processing of increasing amounts of data is a time-consuming task. In this paper, several modifications of methods for the Russian language are proposed. It also includes the results of experiments comparing a keyword extraction method, vocabulary method, and some methods based on neural networks. Text collections for these tasks exist for the English language and are actively used by the scientific community, but at present, such datasets in Russian are not publicly available. In this paper, we present a corpus of scientific texts in Russian, RuSERRC. This dataset consists of 1600 unlabeled documents and 80 labeled with entities and semantic relations (6 relation types were considered). The dataset and models are available at \url{https://github.com/iis-research-team}. We hope they can be useful for research purposes and development of information extraction systems.
\end{abstract}

\begin{IEEEkeywords}
entity recognition, relation classification, neural network models, dataset building, information extraction
\end{IEEEkeywords}

\section{Introduction}
With the spread of the Internet, the amount of information, including textual information, is growing extremely fast. According to the journal “Nature”, the world scientific community publishes annually over a million articles just on biomedical topics \cite{landhuis2016scientific}. Scientific publications contain valuable information on leading scientific advances, but efficiently processing such a huge amount of data is a time-consuming task.

One of the information extraction tasks is Entity Recognition (ER). The goal of this task is entities detection and classification on the predefined categories such as names, organizations, locations, temporal expressions, money, etc. This task is often solved together with the Relation Extraction (RE) task, the essence of which is to find pairs of entities which can be linked by a semantic relation. If a set of relations is predefined, then we talk about the Relation Classification (RC) task - matching each pair of entities with a particular semantic relation. Often the following assumption is made: the entities must be in the same sentence.

Nowadays deep learning based methods quite well solve these tasks. Such methods use language models built on large unlabeled corpora, for example, Wikipedia. To achieve good quality on data from specific domains, one needs to fine-tune models on special corpora. In this paper we describe a process of creating such text collection. Our corpus contains texts on information technology domain and is called the RuSERRC dataset (Russian Scientific Entity Recognition and Relation Extraction). We also conducted several experiments to research and compare different methods on this corpus. The dataset and models are publicly available.

\section{Related Works}
Named entity recognition and relation classification are important steps to extract information from texts. The most famous datasets in English for these tasks are CONLL04 \cite{carreras-marquez-2005-introduction}, TACRED \cite{zhang-etal-2017-position}, SemEval-2010 Task 8 \cite{hendrickx-etal-2010-semeval}. For the Russian language, the most popular one is the FactRuEval-16 dataset \cite{FactRuEval2016}, that consists of news texts.
Currently, methods based on Transformer architecture are considered to be the most promising. Transformers typically go through semi-supervised learning involving unsupervised pre-training followed by supervised fine-tuning for the task under consideration. An example of BERT fine-tuning is presented in \cite{soares2019matching}. To achieve relation representation by fine-tuning BERT with a large scale “matching the blanks” pre-training entity linked texts are used. This method performs well on the SemEval-2010 Task 8 dataset (F1-score of 89.5\%) and outperforms previous methods on TACRED (F1-score of 71.5\%).

In order to achieve good quality on data of specific domain, it is necessary to fine-tune a model on a proper dataset. Pre-training is usually done on a much larger dataset than fine-tuning for a specific knowledge domain, due to the restricted availability of labeled training data. Recently, as a useful testing ground a corpus of Russian strategic planning documents (RuREBus) became available \cite{Artemova2020SoWT}. The BERT-based model obtains the best F1-score of 0.561 for NER and 0.441 for RE on it. Another new dataset in Russian is RURED \cite{davletovetal2020}. It consists of economic news texts. The multilingual BERT model yields 0.85 for NER and the SpanBERT \cite{joshi2020spanbert} model gives the best results of 0.78 for RE F1-score on this corpus.

Solving the tasks of extracting entities and classifying semantic relations in scientific texts is also of interest. Although there are tools (e.g. https://github.com/natasha/natasha, https://github.com/buriy/spacy-ru) for extracting traditional entity types from general domain texts (persons, locations, organizations, etc.), extraction of entities and relations from scientific and technical texts in Russian still needs research. Such collections for the English language \cite{gabor-etal-2018-semeval}, \cite{DSouza2020TheSD}, \cite{smith2019scienceexamcer} are available and actively used by the scientific community, however, at present, we could not find a similar dataset in Russian.

\section{Dataset Annotation}
Collected corpus consists of abstracts of scientific papers from the information technology domain. The texts were taken from journals “Vestnik NSU. Series: Information Technologies”\footnote{\url{https://journals.nsu.ru/jit/archive/}} and “Software \& Systems”\footnote{\url{http://www.swsys.ru/}}. The corpus consists of 1600 unlabeled texts and 80 texts which were annotated manually with terms and semantic relations between them. Each document was annotated by two annotators, all  disagreements were resolved by a moderator. The annotators' agreement in the entity recognition task is 71.57\%. The value is calculated as the ratio of the intersection of the selected terms to the union of the selected terms. The resulting value shows a high degree of subjectivity in finding terms, and in determining the exact boundaries of entities, which indicates the complexity of the problem being solved. The annotators' agreement in the relation classification task is 70.03\%. The value is also calculated as the ratio of the intersection of the selected relations to the union of the selected relations. More detailed description of entities and relations is provided below.

\subsection{Entities}
As entities we consider nouns or noun groups, which are terms in this particular domain. By the term we mean a word of phrase which is name of certain concept of a field of science, technology, art, etc. We consider as entities terms consisting of one token or an abbreviation (\textit{“database” \textcyrillic{(БД)}, “software” \textcyrillic{(ПО)}, "interface" \textcyrillic{(интерфейс)}}), names of programming languages (\textit{“Python”, “Java”, “C++”}) and libraries (\textit{“Pytorch”, “Keras”, “pymorphy2”}), hyphenated concepts containing Latin characters (\textit{"n-gram" \textcyrillic{(n-грамма)}, "web service" \textcyrillic{(web-сервис)}}). Terms with misspellings or misprints were labeled as entities as well. Entities are listed with "," ";" or connected by a conjunction "and" (\textcyrillic{и}) were marked separately, if possible. Moreover, we annotate some abstract concepts (for example, \textit{“method” \textcyrillic{(метод)}, “phenomenon” \textcyrillic{(явление)}, “property” \textcyrillic{(свойство)}}, etc.) as terms to link entities through these concepts with a knowledge base in the future (for example, for special domain ontology creation).

The main difficulty is the process of differentiation between terms and non-terms. It is often difficult to understand without context whether the phrase is a term or not. We consider a multi-word term as a chain of tokens of maximum length, from which a more general term is derived, if tokens are removed. For example, the compound term \textit{“model of structural organization of a single information space” \textcyrillic{(модель структурной организации единого информационного пространства)}} is considered to be a term in the corpus, because a single word \textit{“model” \textcyrillic{(модель)}} does not reflect the exact meaning in this context. But in case when a word \textit{“model” \textcyrillic{(модель)}} appears in the text without additional words, we consider it as a term. Usually, such entities are the names of software products, methods, algorithms, tasks, approaches (\textit{"Android operating system" \textcyrillic{(операционная система Android)}, "k nearest neighbors method" \textcyrillic{(метод k ближайших соседей)}, "support vector machine" \textcyrillic{(метод опорных векторов)}}.

In scientific texts in Russian, verbal nouns are often found that denote processes. From the point of view of semantics, the process leads to changes, affects the result. That is why it is desirable to include such nouns in the entity; it affects the extraction of relations. Examples: \textit{"image processing" \textcyrillic{(обработка изображений)}, "system testing" \textcyrillic{(тестирование системы)}, "text analysis" \textcyrillic{(анализ текстов)}}, etc.

The entities were marked in the BIO format: each token is assigned a B-TERM tag if it is the initial tag for an entity, I-TERM if it is inside a term, or O if it is outside any entity. Entities are not recursive and do not overlap. As a result, we annotated 80 texts which contain 11 157 tokens and 2 027 terms. The averaged term length is 2.43 tokens. The longest term contains 11 tokens.

\subsection{Relations}
The list of relations was selected as a result of the analysis of the papers \cite{hendrickx-etal-2010-semeval}, \cite{gabor-etal-2018-semeval}, \cite{aditya2019uncovering} based on the following criteria. At first, a relation should be monosemantic (for example, we don’t consider semantic relation $<$Entity-Destination$>$ because it has indirect meaning as well). Secondly, a relation should link scientific terms (for example, in relation $<$Communication-Topic$>$ (an act of communication is about topic) the actants are not scientific terms). Thus, six semantic relations were selected.

\texttt{CAUSE} is a relation of causation, $X$ yields $Y$, for example, [\textit{high-energy beam interaction : deformation \textcyrillic{(взаимодействие высокоэнергетичных пучков : деформация)}}].

\texttt{COMPARE} is a relation of comparison, $X$ is compared to $Y$, for example, [\textit{relational databases : object oriented databases \textcyrillic{(реляционные базы данных : объектно-ориентированные базы данных)}}].

\texttt{ISA} is a relation of taxonomy, $X$ is a $Y$, for example, [\textit{Python : programming language \textcyrillic{(Python : язык программирования)}}].

\texttt{PARTOF} is a relation of meronomy, $X$ is a part of $Y$, for example, [\textit{module : system \textcyrillic{(модуль : система)}}].

\texttt{SYNONYMS} is a relations of synonymy, $X$ is a synonym of $Y$, for example, [\textit{GPU : graphics processing unit \textcyrillic{(GPU : графический процессор)}}].

\texttt{USAGE} is a relation of usage, $X$ is used for $Y$, for example, [\textit{statistical processing method : text mining \textcyrillic{(метод статистической обработки : анализ текстов)}}].

Relations between entities were annotated within one sentence. As a result, 620 relations between terms were labelled: \texttt{CAUSE} – 25, \texttt{COMPARE} – 21, \texttt{ISA} – 90, \texttt{PARTOF} – 77, \texttt{SYNONYMS} – 22, \texttt{USAGE} – 385.

\section{Methods}
We conducted a series of experiments as baselines using both transformer architectures and traditional methods for solving the task of term extraction, among them we implemented a dictionary-based method, combined method and statistical method. All methods descriptions are provided below.

\subsection{Dictionary-based Method for Entity Recognition}
This approach proposes to use a predefined set (dictionary) of terms. It was collected semi-automatically in two ways.
\begin{enumerate}
    \item We extracted 2-, 3- and 4-gramms from the scientific papers and sorted them by the TF-IDF value; then we manually filtered phrases, which potentially can be terms.
    \item We extracted all articles titles from Wikipedia, which are included in a subgraph of category “Science”, and then manually selected phrases, which potentially can be terms.
\end{enumerate}
Thus we collected 17252 terms.

\subsection{Combined Method for Entity Recognition}
The main difficulty in conducting experiments with usage of different machine learning algorithms is the lack of labeled data. To solve this problem we automatically annotated 1118 scientific papers (which were cleaned from formula, tables, pictures etc.) with the terms from our dictionary, described above. Thus we got an annotated dataset, which consists of ~2 million tokens and ~177K terms. The input sequence was encoded in chars-level. The model contains one bidirectional LSTM layer and CRF layer to form an output sequence of tags. All metrics are described in Table 1.
We analyzed how many terms the model was able to extract, that were not in the terms dictionary, and found about 26.5\% of all unique terms are phrases that the model has not previously seen.

The RAKE algorithm (Rapid automatic keyword extraction) is well applicable to dynamic corpora and completely new domains, it does not depend on the document language and its features \cite{rose2010automatic}. The first step of the algorithm is applying stop-words and a delimiter list to highlight multi-word terms. After that, some statistical information is calculated. For each word, the frequency with which it occurs is evaluated. Second parameter is a count of relations between current word and other words in the text. Based on these values, the weight of each key phrase is estimated. All phrases are sorted by their weight, so the most likely phrases get the maximum value.

We used the implementation of RAKE, which supports Russian language\footnote{\url{https://github.com/vgrabovets/multi\_rake}} and automatic extraction of stop-words from the text. We noticed that the algorithm often adds key phrases containing verb forms. Since we consider only nouns or groups of nouns as entities, we decided to preprocess texts and remove all verbs and their forms before applying RAKE. Verb forms were extracted using Mystem\footnote{\url{https://github.com/nlpub/pymystem3}}.

\subsection{Entity Recognition and Relation Classification Using BERT-based models}

RuBERT \cite{kuratov2019adaptation} is a BERT model pre-trained on the Russian Wikipedia. Using this model's weights for initialization we pre-trained BERT on our collection of scientific texts.

\textbf{BertLinearER}. The simplest way to mark up sequences with BERT is to use a linear layer on top of the vector representations of tokens generated by BERT. The loss function in this case is based on cross entropy (CrossEntropyLoss).

\textbf{BertLstmLinearER and BertCnnLinearER}. A reasonable solution is to make the classifier more complex by adding layers which learn how to find dependencies in a sequence. Convolutional layers are more suitable for allocating short-term dependencies (for tokens at a small distance from each other) and LSTM (Long Short-term memory) layers are good for allocating long-term dependencies (for tokens at a large distance from each other). The convolutional layers are combined into blocks (CNN blocks) whose structure is based on the Keras-bert-ner project\footnote{\url{https://github.com/liushaoweihua/keras-bert-ner/}}. Loss function is also based on cross entropy.

\textbf{BertLstmCrfER and BertCnnCrfER}. The next step for improving the classifier is to add a CRF layer. Another loss function, the likelihood logarithm, is used along with CRF.

\begin{figure}[htbp]
    \includegraphics[width=\columnwidth]{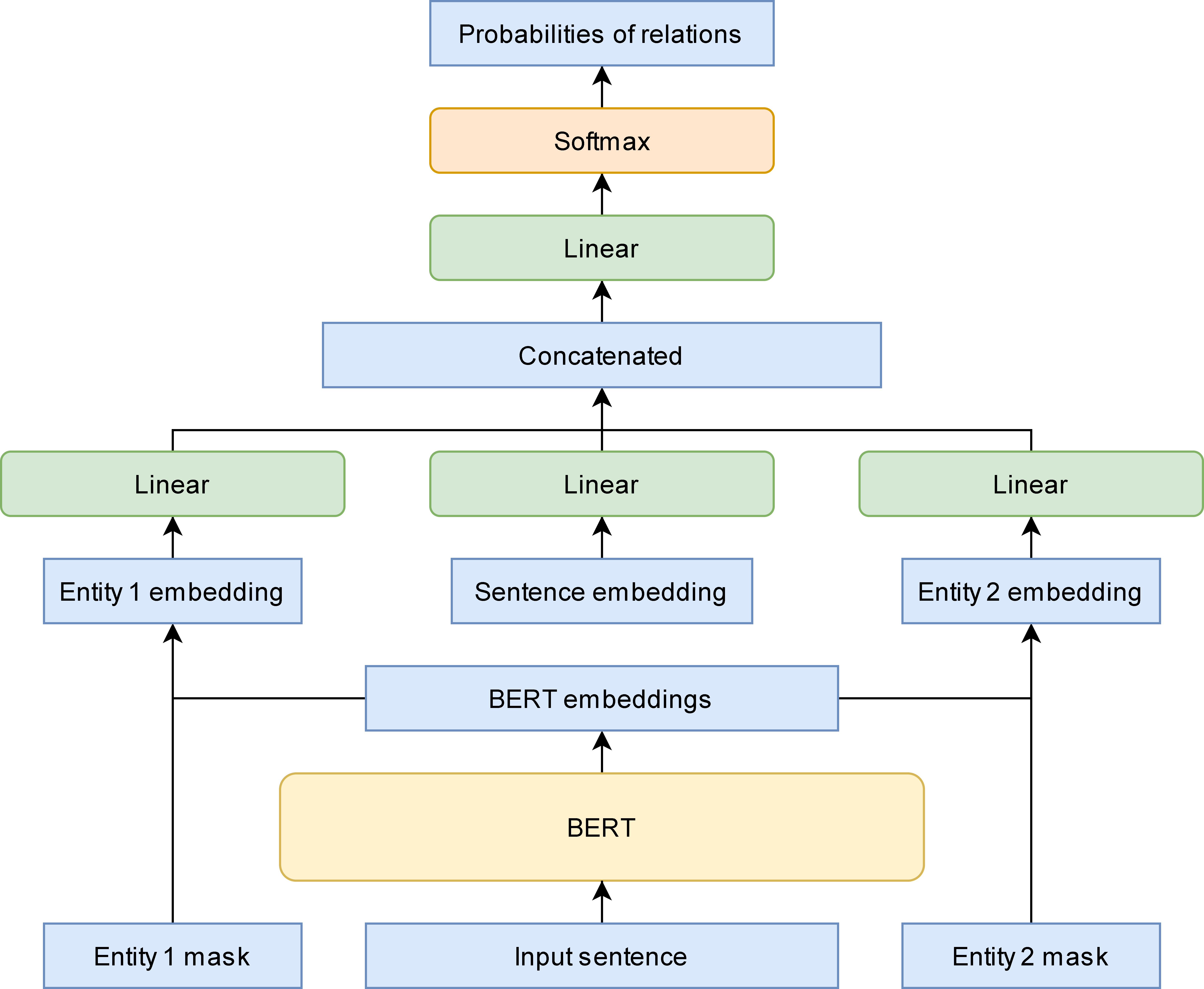}
    \caption{Model architecture for relation classification.}
    \label{fig1}
\end{figure}

\textbf{BertRC}. A baseline for relation classification is based on the R-BERT architecture \cite{wu2019enriching}. The input data for RC is different from NER — in addition to the token sequence, bitmasks are provided for the input. These masks show that tokens belong to entities. The softmax layer is located at the model output. MSE loss is used as the loss function. Thus, we used the following architecture (see Fig.~\ref{fig1}).

\section{Experiments and Results}
\subsection{Comparison of Entity Recognition Algorithms}
To assess the quality of the algorithms for entity recognition we consider both exact-token matching and fuzzy-token matching (see Table~\ref{tab1}). These metrics were calculated on the entire RuSERRC dataset since it was not used during the fitting algorithms.

The low metrics of the dictionary-based method are related to the fact that the dictionary consists only of the full names of the terms, while in the real texts the sequence of term tokens can be broken by other tokens, contain synonyms, abbreviations or even be incomplete. The low metrics of the combined method are due to the same reason, since texts for training were automatically marked up using this dictionary, therefore, in the training set there were no examples in which the term was somehow changed.

The metrics of the RAKE algorithm are slightly better: the algorithm extracts more terms from the text. Optimization with the removal of verb forms reduces the power of the extracted terms set, thereby increasing the accuracy of the algorithm.

\begin{table}[b]
\caption{Tokens Matching.}
\begin{center}
\begin{tabular}{|c|c|c|c|}
\hline
\multicolumn{4}{|c|}{\textbf{Exact-token matching.}} \\
\hline
\textbf{Method} & \textbf{\textit{Precision}}& \textbf{\textit{Recall}}& \textbf{\textit{F-score}} \\
\hline
Dictionary-based & 0.25 & 0.17 & 0.20\\
\hline
Combined & 0.19 & 0.13 & 0.15\\
\hline
RAKE & 0.36 & 0.28 & 0.32\\
\hline
Optimized RAKE & {\bfseries 0.44} & {\bfseries 0.35} & {\bfseries 0.39}\\
\hline
\multicolumn{4}{|c|}{\textbf{Fuzzy-token matching.}} \\
\hline
Dictionary-based & {\bfseries 0.82} & 0.34 & 0.48\\
\hline
Combined & 0.82 & 0.28 & 0.42\\
\hline
RAKE & 0.62 & {\bfseries 0.63} & {\bfseries 0.63}\\
\hline
Optimized RAKE & 0.65 & 0.57 & 0.61\\
\hline
\end{tabular}
\label{tab1}
\end{center}
\end{table}

\subsection{Comparison of BERT-based Models}
All models were trained during 30 epochs (complete passes through the training dataset). 10\% of the dataset was allocated for testing (these examples did not participate in model training). A comparison of F-score on the RuSERRC corpus for the considered BERT-based models for entity recognition is presented in Table ~\ref{tab3}. The BERT-based model for relation classification gives 0.840 of F-score.

\begin{table}[htbp]
\caption{BERT-based models comparison.}
\centering
\label{tab3}
\begin{center}
\begin{tabular}{|l|l|}
\hline
\bfseries Model & \bfseries F-score\\
\hline
LinearER & 0.520\\
\hline
LstmLinearER & {\bfseries 0.530}\\
\hline
CnnLinearER & 0.496\\
\hline
LstmCrfER & 0.522\\
\hline
CnnCrfER & 0.503\\
\hline
\end{tabular}
\end{center}
\end{table}

Table~\ref{tab4} shows that the model recognizes the USAGE relation best, which is not surprising, since it has the highest frequency in the training sample. The results can be improved by expanding the corpus of texts for BERT pre-training and increasing the number of epochs for pre-training and training.

\begin{table}[htbp]
\caption{Metrics for relations classification.}
\centering
\label{tab4}
\begin{center}
\setlength{\tabcolsep}{4pt}
\begin{tabular}{|c|c|} \hline
\textbf{Relation} & \textbf{F-score} \\ \hline
\texttt{CAUSE} & 0.75 \\ \hline
\texttt{COMPARE} & 0.67 \\ \hline
\texttt{ISA} & 0.90 \\ \hline
\texttt{PARTOF} & 0.64 \\ \hline
\texttt{SYNONYMS} & 1.00 \\ \hline
\texttt{USAGE} & 0.88 \\ \hline

\end{tabular}
\end{center}
\end{table}

Due to the fact that we could not find a similar corpus and models for the Russian language for solving the tasks set, the published results of other researchers on other similar datasets aims to show what results the model can achieve in principle. For the ER task, F-score is 0.561 on RuREBus \cite{Artemova2020SoWT}, 0.850 -- on RURED \cite{davletovetal2020}, 0.703 -- on SpERT \cite{eberts2019span}; for the RC task F-score is 0.441 on RuREBus, 0.762 -- on RURED, 0.508 -- on SpERT.
Of course, when analyzing the results, it is important to take into account that the values of the metrics on the dataset largely depend on its properties: size, completeness, quality of texts, training examples and other characteristics.

\section{Conclusion}
This paper describes experiments of comparing various methods for automatic extraction of entities and semantic relation classification for building models that work with Russian language. These models are especially relevant, since most of the existing studies are focused on data in English and Chinese and it’s quite difficult to find high-quality models for Russian in the public domain.

In the future, we plan to conduct a series of experiments with the ERNIE model \cite{zhang2019ernie}, that uses additional structured information about the language. Currently, pre-trained ERNIE models exist for only two languages — English and Chinese. According to the conclusions made in \cite{zhang2019ernie}, there are reasons to believe that such an approach will improve the results for Russian.

The RuSERRC dataset of scientific texts that we built contains markup of entities and semantic relations between them. This corpus is publicly available, and we hope it can be useful for research purposes and developing information extraction systems.




\bibliographystyle{IEEEtran}
\bibliography{IEEEabrv,references}

\end{document}